\documentclass{article}
\usepackage[utf8]{inputenc}
\usepackage{amsmath}
\usepackage{amssymb}
\usepackage{mathtools}
\usepackage[ruled,vlined]{algorithm2e}

\newcommand\ndots{\stackrel{\mathclap{\normalfont\mbox{N}}}{\dots}}

\title{Spectral Convolution Networks}
\author{Maria Francesca \and Arthur Hughes \and David Gregg}
\date{Trinity College, Dublin\\ November 16\textsuperscript{th} 2016}

\begin{document}

\maketitle

\begin{abstract}
Previous research has shown that computation of convolution in the frequency domain provides a significant speedup versus traditional convolution network implementations.
However, this performance increase comes at the expense of repeatedly computing the transform and its inverse in order to apply other network operations such as activation, pooling, and dropout. 
We show, mathematically, how convolution and activation can both be implemented in the frequency domain using either the Fourier or Laplace transformation.
The main contributions are a description of spectral activation under the Fourier transform and a further description of an efficient algorithm for computing both convolution and activation under the Laplace transform. 
By computing both the convolution and activation functions in the frequency domain, we can reduce the number of transforms required, as well as reducing overall complexity.
Our description of a spectral activation function, together with existing spectral analogs of other network functions may then be used to compose a fully spectral implementation of a convolution network.
\end{abstract}

\section{Motivation}
Convolution networks are used for machine learning problems such as image classification, natural language processing, and recommendation systems \cite{mathieu,yann,van}. They are represented as a graph of operators which are typically sequentially applied to some input image, eventually yielding a classification for that input.

Convolution is an expensive operation which is replicated repeatedly within a single network. 
Computation of convolution in the frequency domain under the Fourier transform has been shown to provide a significant speedup versus traditional convolution network implementations \cite{vasilache, mathieu}.
However, typically activation is run following convolution in practise, and previous researchers have been unable to find a spectral implementation for both.
In this paper we describe spectral representations of the max activation function \(a(x)\) paying particular attention to computational complexity. Specifically, \(a(x) = max(0,x)\). The abbreviation ReLU is used throughout to designate the part of the network which computes this function.

The following diagrams show part of a hypothetical convolution network, which computes two convolution operators (denoted \(C\)), each with a following activation (denoted \(A\)). This is similar to GoogLeNet, which has two sequential convolution/activation blocks in each inception module \cite{googlenet}. Superscripts show the complexity of each step. It should be noted that the \(O(n^2)\) complexity of convolution is the worst case complexity, where both images being convolved are of similar size. For a large image of \(n\) pixels and a small kernel of \(k\), the complexity is more correctly \(O(nk)\).

\[\ldots \rightarrow \overset{O(n^2)}{C} \rightarrow \overset{O(n)}{A} \rightarrow \overset{O(n^2)}{C} \rightarrow \overset{O(n)}{A} \rightarrow \ldots\]

Previous researchers have noted that it is possible to reduce the overall complexity of the \(C\) block by computing the Fourier transform of the inputs to that block, multiplying these transformed inputs, then computing the inverse Fourier transform \cite{vasilache, mathieu}. This adaptation reduces overall complexity, but introduces a lot of memory transfer. This previous work yields a diagram with additional operators, \(F\) and \(F^{-1}\), which compute the Fourier transform and it's inverse respectively.

\[\ldots \rightarrow \overset{O(nlogn)}{F} \rightarrow \overset{O(n)}{C} \rightarrow \overset{O(nlogn)}{F^{-1}} \rightarrow \overset{O(n)}{A} \rightarrow \overset{O(nlogn)}{F} \rightarrow \overset{O(n)}{C} \rightarrow \overset{O(nlogn)}{F^{-1}} \rightarrow \overset{O(n)}{A} \rightarrow \ldots\]

As the diagram shows, the increase in performance comes at the expense of repeatedly computing the transform and its inverse in order to apply other network operations, including activation, pooling, and dropout (though activation is the most common). The next step seems to be a fully spectral convolution network, or at least one which is computed in the frequency domain in larger block, so as to avoid these constant transforms.

Recent work has addressed the idea of spectral pooling \cite{rippel}, but activation remains an open issue. Multiple articles on the topic have cited the lack of appropriate representation of many common activation functions in the frequency domain to be a significant block to future work in this area \cite{rippel,mathieu,vasilache}.

We show, mathematically, how convolution and activation can both be implemented in the frequency domain during inference. We also explain how to implement activation during training, but note some problems which may make this inadvisable. Our proposal is a representation which yields the following diagram for the case above during inference:

\[\ldots \rightarrow \overset{O(nlogn)}{F} \rightarrow \overset{O(n)}{C} \rightarrow \overset{O(n)}{A} \rightarrow \overset{O(n)}{C} \rightarrow \overset{O(n)}{A} \rightarrow \overset{O(nlogn)}{F^{-1}} \rightarrow \ldots\]

By applying the activation function in the frequency domain, we reduce the number of Fourier transforms to only two at each boundary between non-spectral parts of the network. Most importantly, this removes the requirement to swap frequently between representations, and so should reduces the amount of memory required to run spectral convolution and activation. Our description of a spectral activation function, together with existing spectral analogs of other network functions may then be used to compose a fully spectral implementation of a convolution network.

% They want us: 
% \cite{mathieu}
% \cite{rippel}
% \cite{vasilache}
% http://arxiv.org/abs/1412.7580

% Convolution in FFT on GPUs: Mathieu, Henaff, LeCun.

% fbfft and the better one.

% Model and Train Except for ReLU: Rippel, Snoek and Adams.

The main contributions are a description of spectral activation under the Fourier transform and a further description of an efficient algorithm for computing both convolution and activation during inference under the Laplace transform. 
These are followed by brief descriptions of previous work on spectral pooling \cite{rippel}, and the fully connected layer, which are included as an aid to others wishing to build on our work.

\section{Spectral Representations of Operations}
In this section we describe the spectral representations of the convolution and activation functions under the Fourier transform, as they are used in a convolution network. Each operation is defined, and the transforms described. 

It is important to remember that the Fourier transform for a function exists only when 1) the integral of the absolute value of the function from \(-\infty\) to \(+\infty\) exists (the limit has some non-infinite value) and 2) any discontinuities in the function are finite \cite{bracewell}. Proving that this holds for activation functions is a key contribution of this paper, as this seems to have been a stumbling point for previous work.

\subsection{Convolution}
Convolution is typically defined as a function from \(\mathbb{R} \to \mathbb{R}\) of two one-dimensional functions, \(f(x)\) and \(g(x)\):

\begin{equation}
f*g = \int_{-\infty}^{\infty}f(\tau)g(x-\tau)d\tau
\end{equation}

In computer science papers it is convenient to state that this function is applied to each pixel in some input image. However, in order to make composition of functions simple, we consider multidimensional convolution, specifically from \(\mathbb{R}^2 \to \mathbb{R}^2\) \cite{bracewell}.

\subsubsection{Multidimensional Convolution}
In general, multidimensional convolution is defined for two n-dimensional function \(f(x_1, x_2, \dots, x_N)\) and \(g(x_1, x_2, \dots, x_N)\):

\begin{equation}
f*\ndots*g = \int_{-\infty}^{\infty} \ndots \int_{-\infty}^{\infty} 
f(\tau_1, \dots, \tau_N)g(x_1-\tau_1, \dots, x_N-\tau_N)
d\tau_1, \dots, d\tau_N
\end{equation}

Since convolution networks typically operate on 2-dimensional images, we will mainly focus on 2-dimensional convolution, operating on functions \(f(x,y)\) and \(g(x,y)\):

\begin{equation}
f**g = \int_{-\infty}^{\infty}\int_{-\infty}^{\infty} 
f(\tau_x, \tau_y)g(x-\tau_x, y-\tau_y)
d\tau_x, d\tau_y
\end{equation}

The convolution theorem also holds for multidimensional convolution \cite{bracewell}. So we know, taking \(\mathcal{F}(f)\) as the Fourier representation of the function f, that:

\begin{equation}
f**g = \mathcal{F}(f)\mathcal{F}(g)
\end{equation}

This is the property that has allowed previous work to increase the speed of computation for convolution neural networks \cite{mathieu, rippel, vasilache}. Although \(f\) and \(g\) are given as functions here, it may be more familiar to think of them directly as images or matrices, and their Fourier representations to be the Discrete Fourier Transform (DFT) of those structures. The parameters to both functions can be thought of as accessing the colour value at some point. Although in practice we will usually use integers to access a floating point representation, these functions should be thought of a mapping from \(\mathbb{R}^2 \to \mathbb{R}\).

It is also important to note that multidimensional convolution includes in its definition an explicit choice of corner-case. In general, the edges of the image being convolved are dealt with using one of three approaches: 1) pixels outside the image are considered zeroed or black (cropping), 2) pixels outside the image are considered to have the value of their closest edge pixel (wrapping), or 3) pixels outside the image are considered to have the value of a mirrored pixel (mirroring). 
It is not typical to define an arbitrary function so that it exhibits wrapping or mirroring, and so 
multidimensional convolution, as stated here, explicitly selects the first approach.
Practitioners may wish to define their image as including a border or interpolated or mirrored pixels in order to circumvent this.

\subsubsection{Multichannel Multidimensional Convolution}
The previous section defines 2-dimensional convolution for a single channel. However, in modern architectures such as GoogLeNet\cite{googlenet} and AlexNet\cite{alexnet}, convolution is applied to multiple channels, and the results are then added before being fed into a rectified linear unit. For some input image \(I\), and set of weight matrices \(W\) composed of elements \(w_i\), this can be written:

\begin{equation}
\sum_{i=1}^{|W|} I \text{**} w_i
\end{equation}

Because the above is a sum, and the Fourier transform is a linear transformation where \(f(x+y) = f(x) + f(y)\), it is clear no further transforms are required for multichannel convolution.

\subsection{Activation}
Rectified Linear Units (ReLU) are components of a network that compute an activation function. These are the functions that introduce non-linearity into the system. In this paper, we will consider the activation function usually defined as \(a(x) = max(0,x)\).

An alternative way to represent \(a\) is by using the Heaviside function \cite{bracewell}.
The Heaviside function, \(H(x)\) is usually described in one dimension as:

\begin{equation}
H(x) = 
    \begin{cases}
        1   &   x > 0   \\
        0   &   x < 0
    \end{cases}
\end{equation}

\noindent Using Heaviside, we can also say that \(a(x) = xH(x)\). This representation is less common in machine learning literature, but is often more convenient when computing integrals, which is required to check that the Fourier transform exists.

% Note here how things change for other activation functions; ESP softmax

However, because we earlier described convolution as a function from \(\mathbb{R}^2 \to \mathbb{R}^2\), and we want to compose these functions, we need to define \(a\) as mapping from \(\mathbb{R}^2\). Since activation does not change the shape of the input layer, this means activation can be written as another function from \(\mathbb{R}^2 \to \mathbb{R}^2\):

\begin{equation}
a(x,y) = max(0, f(x,y))
\end{equation}

\noindent Similarly, we can also model our activation function \(a\) using a modified Heaviside function also defined for \(\mathbb{R}^2\):

\begin{equation}
H(x,y) = 
    \begin{cases}
        1   &   x>0 \wedge y>0  \\
        0   &   x\le0 \vee y\le0
    \end{cases}
\end{equation}

\noindent Using \(H\), the activation function \(a\) can be re-stated:

\begin{equation}
a(x,y) = f(x,y)H(f(x,y)) = 
    \begin{cases}
        f(x,y)   &   x>0 \wedge y>0  \\
        0       &   x\le0 \vee y\le0
    \end{cases}
\end{equation}

Having described an example activation function, we  must check whether it has a representation in the frequency domain. This means that we need to make sure that any discontinuities in \(a\) are finite, and that the integral from \(-\infty\) to \(\infty\) exists. 

While there is only one point (at 0) where \(a\) is discontinuous, the integral from \(-\infty\) to \(\infty\) yields infinity, so \(\mathcal{F}(f(x,y)H(x,y))\) does not exist:

\begin{equation}
\begin{aligned}
\mathcal{F}(a)
&=  \int_{-\infty}^{\infty}\int_{-\infty}^{\infty}f(x,y)H(x,y) dx dy\\
&= f(x,y)\int_{-\infty}^{\infty}\int_{-\infty}^{\infty} H(x,y) dx dy\\ 
&= \infty
\end{aligned}
\end{equation}

In general, this will also be true for many activation functions commonly used to construct convolution networks, as the role of the ReLU is to activate values from the underlying layer.
When (in theory) there is an infinitely large value in the underlying layer, \(a(\infty)\) will also often tend to infinity. This means the integral of \(a\) will also tend to infinity, and imply that we cannot compute a transform to the frequency domain for that function.
This problem is solved by considering the support of the activation function, especially in our use case; composition with convolution.

\subsection {Activation \(\circ\) Convolution}
For both single channel and multichannel convolution in a network, their composition with an activation function is of the same shape. For single layer convolution of an image \(I\) with a kernel \(w\), we have:

\begin{equation}
(a \circ **)(I, w)
\end{equation}

\noindent For multilayer convolution with a set of kernels \(W\), the composition is:

\begin{equation}
(a \circ \sum_{i=1}^{|W|} **)(I, W)
\end{equation}

Either case may be abstracted, by defining a function \(c(x,y)\) which represents either type of convolution, and always specifying that the weight matrices are in a set \(W\). For single channel convolution this set simply contains only one element. The function \(c(x,y)\) is defined to be either single or multichannel convolution, depending on the network architecture. The abstract composition looks like:

\begin{equation}
(a \circ c)(I,W)
\end{equation}

It is important to note that \(c\) has finite support. This means that there are a finite set of values for which \(c\) is non-zero, and is true of convolution where the input functions are both functions with finite support, which is clearly the case in convolution networks, as images and weight matrices have finite size. We also know that \(c\)'s support is exactly known, as well as being finite, due to the Titchmarsh convolution theorem \cite{titchmarsh}.

Since \(c\) has finite support, \(a \circ c\) also has finite support, at most equal to the support of \(c\). This should improve our ability to compute the Fourier transform of the composition, since the composition will be zero almost everywhere, and so we will not run into the problems involved with infinite integrals that we saw earlier.

\subsubsection{Finitely Supported Activation}
It is not practical for an infinitely large image or matrices to be stored in computer memory, however our activation function, \(a\), defined earlier has no upper bounds. Currently, its support is:

\begin{equation}
supp(a) = \{(x,y) | x > 0 \wedge y > 0\}
\end{equation}

\noindent but \(f(x,y)\) is only non-zero within the valid values of the image or weight matrix, so in practice we have a finite-support activation function \(a_{finite}(f(x,y))\):

\begin{equation}
\begin{aligned}
a_{finite}(x,y) &=
    \begin{cases}
        f(x,y)   &   p>x>0 \wedge q>y>0  \\
        0       &   x\le0 \vee y\le0
    \end{cases}\\
    %&= (x,y)(H(x,y) - H(x-p, y-q))
\end{aligned}
\end{equation}

\noindent and its support:

\begin{equation}
supp(a_{finite})= \{(x,y) | p>x>0 \wedge q>y>0\}
\end{equation}

The two new variables in this equation, \(p\) and \(q\), can be thought of as the height and width of the image to which activation is being applied, or equally, as the maximum \((x,y)\) where \(c\) has support.

With these bounds in place we can return to the problem of finding the integral from \(-\infty\) to \(\infty\). If this integral exists, then the Fourier transform of the activation function exists, and we can compute multichannel convolution in the frequency domain.

As before, we model our activation function as multiplication by a modified Heaviside function, and so the integral becomes:

\begin{equation}
f(x,y)\int_{0}^{p}\int_{0}^{q} H(x,y) dx dy
\end{equation}

Since \(H(x,y)\) is always equal to 1 as \(x\) and \(y\) vary between 0 and \(p\) or \(q\) respectively, the result of this integral is some constant times \((x,y)\).
So long as the image contains no infinite values at any point, then \((x,y)\) is also finite, the Fourier transform exists, and can be stated:

\begin{equation}
\begin{aligned}
\mathcal{F}(u,v)    &= \int_{-\infty}^{\infty}\int_{-\infty}^{\infty} a(x,y)e^{-j2\pi(ux+vy)} dx dy,\\
a(x,y)              &= \int_{-\infty}^{\infty}\int_{-\infty}^{\infty} \mathcal{F}(u,v)e^{j2\pi(ux+vy)} du dv
\end{aligned}
\end{equation}

\noindent Because of the existence of this transform, we know that we can compute the activation function in the frequency domain, but there are concerns about how to do so efficiently.

\subsection{Computing Activation Efficiently}

While it is encouraging that we have the mathematical background required for the activation function to exist in the frequency domain, because of the double-integral, it seems we have merely swapped expensive convolution for expensive activation. In fact, we have only moved the convolution to a different stage of computation, as integration can be computed by convolution with the previously-mentioned Heaviside function \cite{bracewell}. However, another option is to consider other transforms to the frequency domain. 

The Laplace transform is another transform to the frequency domain. 
As the Discrete-Time Fourier transform computed using FFT is the discrete version of the Fourier transform, the discrete version of the Laplace transform is the Z transform. 
In this section we provide mathematical justification for using the Laplace transform to compute both convolution and activation in the frequency domain. The Laplace transform of a function \(f(t)\) is defined:

\begin{equation}
\mathcal{L}(f(s)) = \int_{0}^{\infty} f(t) e^{-st} dt
\end{equation}

Convolution under the Laplace transform still obeys the convolution theorem \cite{bracewell}. That is, for an image \(I\) and weight matrix \(w\), convolution is defined:

\begin{equation}
I \text{**} w = \mathcal{L}(I)\mathcal{L}(w)
\end{equation}

As we saw previously, the max activation function can be represented either as a piecewise function, or as multiplication by a Heaviside function. This function is also commonly described as a ramp function.
Both the ramp and Heaviside functions have a known transform under the Laplace transform, making activation simple to compute.
We choose to multiply by Heaviside, as multiplication is simpler than application of a function.
Multiplication of two series under the Laplace transform is performed by computing an integral. The equation for the multiplication of functions \(f(x)\) and \(g(x)\) under the Laplace transform is:

\begin{equation}
\mathcal{L}(f(x)g(x)) = \frac{2}{2 \pi i} \lim_{T \to \infty} \int_{c-iT}^{c+iT} \mathcal{L}(f(\sigma))\mathcal{L}(g(s+\sigma)) d\sigma
\end{equation}

\noindent In our case, the functions \(f(x)\) and \(g(x)\) will be the convolved ``image" (listed as \(c\) in equation 13), and a discrete representation of the Heaviside, or step function. The Heaviside function under the Laplace transform is given:

\begin{equation}
\mathcal{L}(H(x)) = \frac{1}{s}
\end{equation}

\noindent
This means that, with \(c(x)\) as the samples from our image after convolution, we can apply activation by computing:

\begin{equation}
\mathcal{L}(c(x)H(x)) = \frac{2}{2 \pi i} \lim_{T \to \infty} \int_{c-iT}^{c+iT} \frac{\mathcal{L}(c(\sigma))}
{s+\sigma}
d\sigma
\end{equation}

We can now compute both steps, activation and convolution, in the frequency domain efficiently. 
The algorithm for this is listed here as Algorithm \ref{alg:spectral_coa}. 
This computation is run both during the forward pass of Propagation, and when analyzing new samples with a trained network.

\begin{algorithm}[H]
  \label{alg:spectral_coa}
  \SetKwInOut{Input}{Input}
  \SetKwInOut{Output}{Output}
 
  \Input{Map \(x_{in} \in \mathbb{R}^{H \times W} \) \\ Filters \(w_i \in \mathbb{R}^{N \times M}\)}
  \Output{Map \(x_{out} \in \mathbb{R}^{H \times W} \) }
  \(c \leftarrow \mathcal{L}(x_{in})\)\;
  \For{\(\forall i\)}{
    \(w \leftarrow \mathcal{L}(w_i)\)\;
    \(c \leftarrow x \odot w\)\;
    \(c \leftarrow L\_MULTIPLY(c, H)\)\;
  }
  \(x_{out} \leftarrow  \mathcal{L}^{-1}(c)\)\;
  \caption{Spectral Convolution with Activation}
 \end{algorithm}
 
Each step in algorithm \ref{alg:spectral_coa} is linear in the number of pixels in the image except for the Laplace transforms, indicated \(\mathcal{L}\), which are log-linear. 
Additionally, only a final, single, inverse Laplace transform is required. 
Although each weight matrix appears to require an additional Laplace transform, in practice, there is no reason these could not be pre-computed. 
Our spectral Laplace activation then, reduces the number of transforms to only two for each sequence of convolution layers.
Further reductions could be made by considering additional spectral operations.

\subsection{Pooling}

Pooling is used in convolution networks to reduce information between computational layers. Exactly performing existing polling (max, average, etc) still requires a transform back to the time domain. However, very interesting work as been done on spectral pooling. This work has been primarilly done for the Time-discrete Fourier Transform, but the ideas should work for any transform.

\subsubsection{Spectral}
An alternative to exactly mapping the existing pooling functions to the frequency domain is to perform Spectral Pooling \cite{rippel}. Spectral Pooling is an alternative to pooling performed entirely in the frequency domain proposed by researchers at MIT and Harvard; it performs dimensionality reduction by truncating the representation in the frequency domain, rather than apply some function such as \(min()\) or \(max()\).  This approach preserves more information per parameter than other pooling strategies
and enables flexibility in the choice of pooling output dimensionality, but is not yet commonly in use, due to the previous lack of a suitable activation function representation.

\subsection{Fully Connected Layer}
With the basic convolution network operations defined, several convolution architectures can now be implemented almost entirely in the frequency domain, reducing both the complexity of convolution and the number of transforms required. 
However, it does not seem possible to run a fully connected layer in the frequency domain.
Therefore, unless you choose to train and output entirely in the frequency domain, the fully connected layer of a spectral convolution network is identical to any non-spectral one. 

\section{Discussion}
In this paper we described how to calculate a spectral activation function as well as previously known spectral representations of convolution and pooling. We also noted how to avoid problems where activation functions tend to infinity, by considering finite support.
This has solved the problem of a lack of ReLU operator for researchers working on spectral representations of neural network.

Our contribution will unblock the research of others, allowing more work to progress on spectral convolution networks, especially during inference \cite{rippel,vasilache}. This is more and more important as large companies invest in spectral convolution to reduce complexity and energy cost\cite{mathieu}. 

Moving forward, we intend to implement these ideas in a full network, and improve performance during training as well as inference.

\section*{Acknowledgements}
This research was supported by Science Foundation Ireland grant 12/IA/1381.

% \subsection{Future Work}
% %\subsubsection{Additional Operators}
% Moving forward, we intend to implement these ideas in a full network. This will also necessitate future theoretical work.
% In particular, it would be interesting to investigate strided filters. A strided filter is a convolution where convolution is only applied to pixels at a set (strided) offset from an initial point. While it can be computed simply in the time domain, the strides ensure that the convolution theorem does not apply. 

% We would also like to work on more more generalised proofs about spectral convolutions. It is likely that using the more general theory of distributions, more elegant and efficient representations could be described \cite{strichartz}.

%\subsubsection{Architectural Improvements -vs- Translation}
%Spectral domain seems to be better for some things like the smallness of weights; \cite{rippel} We should look into that.

\bibliographystyle{acm}

\begin{thebibliography}{1}

\bibitem{bracewell}
{\sc Bracewell, R.~N.}
\newblock {\em The Fourier transform and its applications}.
\newblock McGraw-Hill, 1976.

\bibitem{alexnet}
{\sc Krizhevsky, A., Sutskever, I., and Hinton, G.~E.}
\newblock Imagenet classification with deep convolutional neural networks.
\newblock In {\em Advances in neural information processing systems\/} (2012),
  pp.~1097--1105.

\bibitem{yann}
{\sc LeCun, Y., Kavukcuoglu, K., Farabet, C., et~al.}
\newblock Convolutional networks and applications in vision.
\newblock In {\em ISCAS\/} (2010), pp.~253--256.

\bibitem{mathieu}
{\sc Mathieu, M., Henaff, M., and LeCun, Y.}
\newblock Fast training of convolutional networks through ffts.
\newblock {\em CoRR abs/1312.5851\/} (2013).

\bibitem{rippel}
{\sc {Rippel}, O., {Snoek}, J., and {Adams}, R.~P.}
\newblock {Spectral Representations for Convolutional Neural Networks}.
\newblock {\em ArXiv e-prints\/} (June 2015).

\bibitem{googlenet}
{\sc Szegedy, C., Liu, W., Jia, Y., Sermanet, P., Reed, S.~E., Anguelov, D.,
  Erhan, D., Vanhoucke, V., and Rabinovich, A.}
\newblock Going deeper with convolutions.
\newblock {\em CoRR abs/1409.4842\/} (2014).

\bibitem{titchmarsh}
{\sc Titchmarsh, E.~C.}
\newblock The zeros of certain integral functions.
\newblock {\em Proceedings of the London Mathematical Society s2-25}, 1 (1926),
  283--302.

\bibitem{van}
{\sc Van~den Oord, A., Dieleman, S., and Schrauwen, B.}
\newblock Deep content-based music recommendation.
\newblock In {\em Advances in Neural Information Processing Systems\/} (2013),
  pp.~2643--2651.

\bibitem{vasilache}
{\sc Vasilache, N., Johnson, J., Mathieu, M., Chintala, S., Piantino, S., and
  LeCun, Y.}
\newblock Fast convolutional nets with fbfft: {A} {GPU} performance evaluation.
\newblock {\em CoRR abs/1412.7580\/} (2014).

\end{thebibliography}

\end{document}